%% file: bmvc_final.tex
\documentclass{bmvc2k}

\usepackage{multirow,color,colortbl,soul,bm,float,booktabs,amssymb,amsmath,amsfonts,wrapfig}
\newcommand{\ie}{\textit{i.e.,} }
\definecolor{softred}{rgb}{1.0,0.6,0.6}
\definecolor{softorange}{rgb}{1.0,0.8,0.6}
\definecolor{softyellow}{rgb}{1.0,1.0,0.6}

\def\ours{$\mathbf{C}^{3}$-GS }
\def\oursnospace{$\mathbf{C}^{3}$-GS}
\def\mdone{Coordinate-Guided Attention }
\def\mdonenospace{Coordinate-Guided Attention}
\def\mdtwo{Cross-Dimensional Attention }
\def\mdtwonospace{Cross-Dimensional Attention}
\def\mdthree{Cross-Scale Fusion }
\def\mdthreenospace{Cross-Scale Fusion}

\title{$\mathbf{C}^{3}$-GS: Learning Context-aware, Cross-dimension, Cross-scale Feature for Generalizable Gaussian Splatting}

\addauthor{Yuxi Hu}{yuxi.hu@tugraz.at}{1}
\addauthor{Jun Zhang}{jun.zhang@tugraz.at}{1}
\addauthor{Kuangyi Chen}{kuangyi.chen@tugraz.at}{1}
\addauthor{Zhe Zhang}{doublez@stu.pku.edu.cn}{2}
\addauthor{Friedrich Fraundorfer}{friedrich.fraundorfer@tugraz.at}{1}

\addinstitution{
 Institute of Visual Computing,\\
 Graz University of Technology,\\
 Graz, Austria
}
\addinstitution{
 School of Electronic and Computer Engineering,\\
 Peking University,\\
 Beijing, China
}

\runninghead{Hu et al.}{$\mathbf{C}^{3}$-GS}

\begin{document}
\maketitle

\begin{abstract}
\input{sec/0_abstract.tex}
\end{abstract}

\input{sec/1_introduction.tex}
\input{sec/2_related_works.tex}
\input{sec/3_method.tex}
\input{sec/4_experiments.tex}
\input{sec/5_conclusion.tex}

\vspace{6pt}
\noindent\textbf{Acknowledgements}
This work was supported by the China Scholarship Council under Grant No. 202208440157.

\bibliography{egbib}

\input{sec/X_suppl}

\end{document}

%% file: sec/0_abstract.tex
Generalizable Gaussian Splatting aims to synthesize novel views for unseen scenes without per-scene optimization. In particular, recent advancements utilize feed-forward networks to predict per-pixel Gaussian parameters, enabling high-quality synthesis from sparse input views. However, existing approaches fall short in encoding discriminative, multi-view consistent features for Gaussian predictions, which struggle to construct accurate geometry with sparse views. To address this, we propose $\mathbf{C}^{3}$-GS, a framework that enhances feature learning by incorporating context-aware, cross-dimension, and cross-scale constraints. Our architecture integrates three lightweight modules into a unified rendering pipeline, improving feature fusion and enabling photorealistic synthesis without requiring additional supervision. Extensive experiments on benchmark datasets validate that $\mathbf{C}^{3}$-GS achieves state-of-the-art rendering quality and generalization ability. Code is available at: \url{https://github.com/YuhsiHu/C3-GS}.

%% file: sec/1_introduction.tex
\begin{figure}[htb]
  \centering
    \begin{subfigure}
      \centering
      \includegraphics[width=0.48\linewidth]{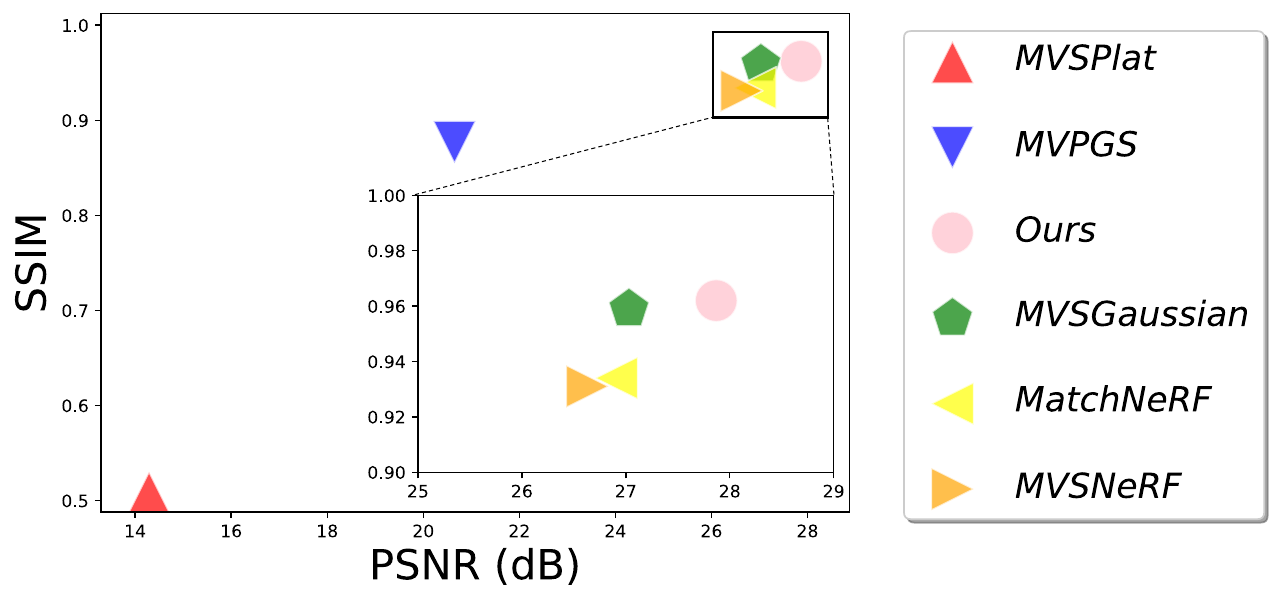}
    \end{subfigure}
    \begin{subfigure}
      \centering
      \includegraphics[width=0.48\linewidth]{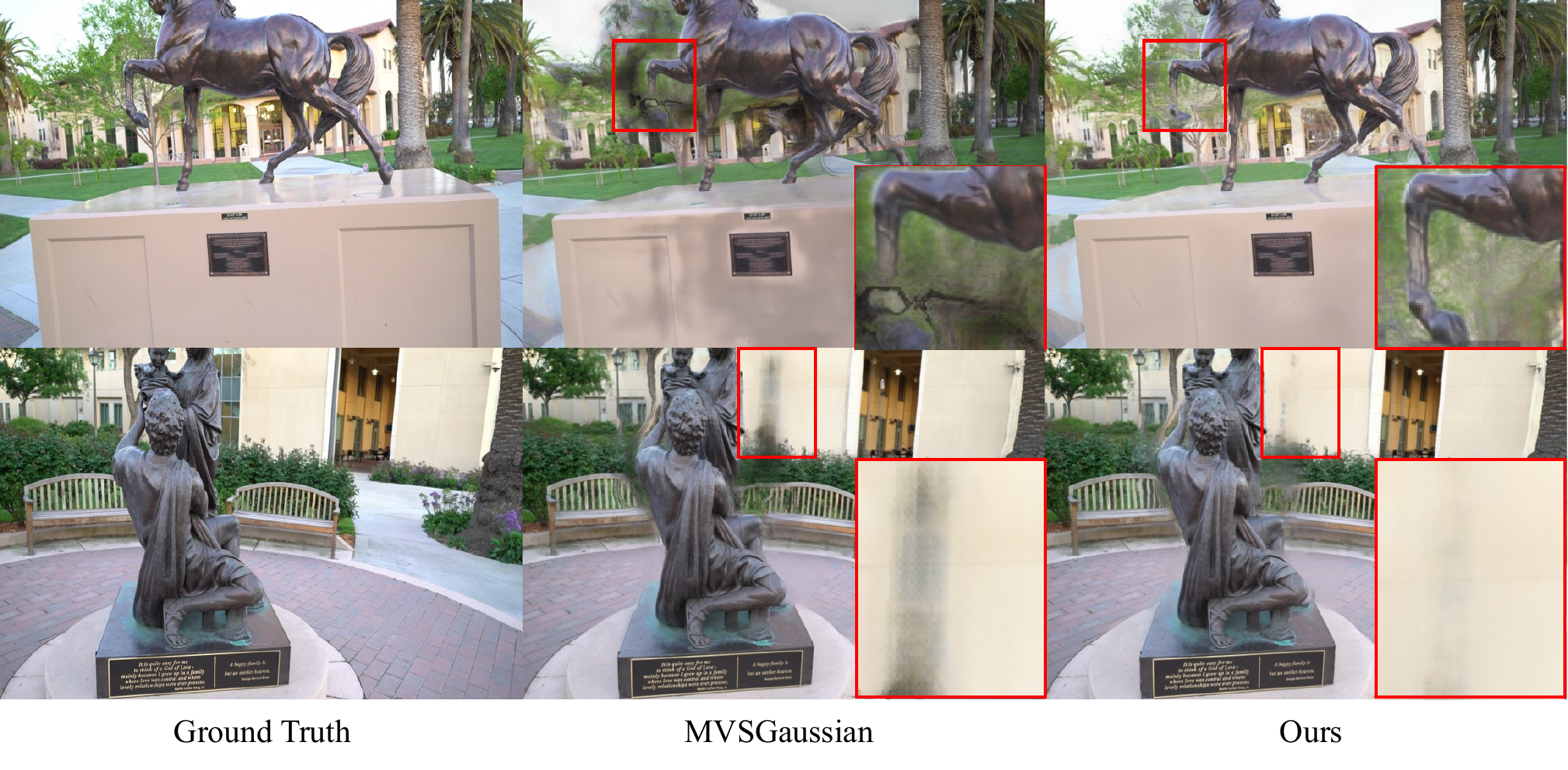}
    \end{subfigure}
    \vspace{-5pt}
    \caption{\textbf{Comparison with existing methods.} \textbf{Left:} Generalization results on DTU~\cite{dtu} with 3 input views, where our method achieves higher PSNR and SSIM. \textbf{Right:} Synthesized images on ``Horse" and ``Family" scenes from Tanks and Temples~\cite{tanks}, highlighting improved visual fidelity.}
    \vspace{-10pt}
  \label{fig:fig1}
\end{figure}

\section{Introduction}
\label{sec:intro}
Novel view synthesis is intended to generate realistic images from novel viewpoints given a series of posed images. It is widely used in applications such as AR/VR, digital content creation, and autonomous driving~\cite{hu20223d,cao2024slcf,cao2024diffssc,chen2025evloc}. Neural Radiance Fields (NeRF)~\cite{nerf} achieve impressive results with implicit MLPs but suffer from slow rendering due to dense point querying. 3D Gaussian Splatting (3D-GS)~\cite{3Dgaussians} replaces implicit fields with explicit Gaussians, enabling real-time synthesis via efficient splatting. Although standard 3D-GS achieves high-quality, photorealistic rendering, it requires dense input views and lengthy per-scene optimization.

Recent developments include generalizable methods~\cite{splatterimage,gpsgaussian,pixelsplat,mvsplat,mvpgs,mvsgaussian}, which are inspired by generalizable NeRF~\cite{pixelnerf,mvsnerf,matchnerf}, and aim to predict Gaussian parameters via a feed-forward network without further optimization. Early approaches~\cite{pixelsplat,mvsplat} relied on two-view feature matching, while later works~\cite{mvpgs,mvsgaussian} incorporated multi-view stereo (MVS) cues to better exploit multi-view geometry. Despite demonstrable progress, challenges still remain in fully utilizing multi-scale feature information for consistent sparse-view geometry constraint, which is essential for predicting accurate Gaussian parameters for high-fidelity rendering.
To this end, we propose \oursnospace, a generalizable GS framework that enhances feature learning across coordinate, spatial, and scale dimensions. Specifically, building upon the baseline MVSGaussian~\cite{mvsgaussian}, we introduce three novel lightweight modules: (1) \mdone(CGA) that embeds coordinate-aware context into 2D features for robust cost matching; (2) \mdtwo(CDA) that constructs 3D spatially-aware descriptors by fusing 2D features with 3D volumes; and (3) \mdthree(CSF) that refines Gaussian opacity across scales to preserve both global structure and fine details.

We validate \ours on four benchmarking datasets of varying scenarios, demonstrating better rendering performance and generalization capability over the state-of-the-art generalizable rendering methods, as showcased in Fig.~\ref{fig:fig1}. Our method excels in challenging cases by capturing fine structures (e.g., horse legs in Fig.~\ref{fig:fig1} (right)) without artifacts.  Additionally, our \ours is capable of predicting more accurate depth maps against the state-of-the-art generalizable methods (Fig.~\ref{fig:depth_vis}, Table~\ref{Tab:depth}), thanks to our enhanced feature learning modules. 

%% file: sec/2_related_works.tex
\section{Related Works}
\label{sec:related works}

\noindent \textbf{Multi-View Stereo (MVS)} aims to reconstruct 3D scenes from multiple overlapping images. Traditional methods~\cite{fua1995object,gipuma,colmap,schonberger2016pixelwise} rely on rule-based features and struggle in complex scenes. Learning-based approaches such as MVSNet~\cite{yao2018mvsnet} build cost volumes from deep features, while cascade frameworks~\cite{gu2020cascade,yan2020dense,yu2020fast,zhang2023geomvsnet} refine depth progressively. Recent transformer-based methods~\cite{ding2022transmvsnet,caomvsformer,li2025srkd,li2025frequency} further capture long-range spatial dependencies. In this paper, we build on a novel view synthesis framework~\cite{mvsgaussian} that combines a cascade MVS architecture with 3D Gaussian Splatting to generate features and depth information.

\noindent \textbf{Neural Radiance Fields (NeRF)} represent scenes as continuous color and density fields using MLPs~\cite{nerf}, achieving high-quality view synthesis but requiring costly per-scene training. Generalizable NeRF methods~\cite{pixelnerf,mvsnerf,matchnerf} address this by learning feature encoders for 3D points, which are decoded into radiance and density. In particular, MatchNeRF~\cite{matchnerf} incorporates feature matching to achieve accurate scene representation. \cite{mvsnerf,liu2024gefu} leverage cost volume to better capture multi-view geometry of the scene. Despite their appreciable success, achieving a balance between rendering quality, speed, and generalization remains challenging.

\noindent \textbf{3D Gaussian Splatting (GS)} represents scenes with anisotropic Gaussians, enabling real-time rendering via differentiable rasterization, avoiding dense points querying as in NeRF, while still requiring dense input data for per-scene optimization. Inspired by generalizable NeRFs, recent works propose to predict Gaussian parameters in a feed-forward manner. For instance,  PixelSplat~\cite{pixelsplat} estimates Gaussians from two views using an epipolar Transformer, which is extended by MVSplat~\cite{mvsplat} to multi-view with cost volume representation. MVPGS~\cite{mvpgs} incorporates pre-trained depth estimation models to enhance supervision. Meanwhile, MVSGaussian~\cite{mvsgaussian} introduces a hybrid rasterization-volumetric rendering framework, achieving superior performance in rendering quality and efficiency. Despite the success of incorporating multi-view constraints for generalizable rendering, existing methods overlook the enhancement of encoded features in exploiting the intra- and inter-view constraints, resulting in limited generalizability when applied across different datasets. Besides that, the reliance on fixed-view settings and extensive supervision often restricts their flexibility and adaptation to new scenes with varying view counts and configurations. To address these limitations, we extend a generalizable 3D-GS framework~\cite{mvsgaussian} with three lightweight modules that enhance feature aggregation and multi-view reasoning, leading to better generalization without additional supervision.

%% file: sec/3_method.tex
\begin{figure}[htb]
    \centering
    \includegraphics[width=0.96\linewidth]{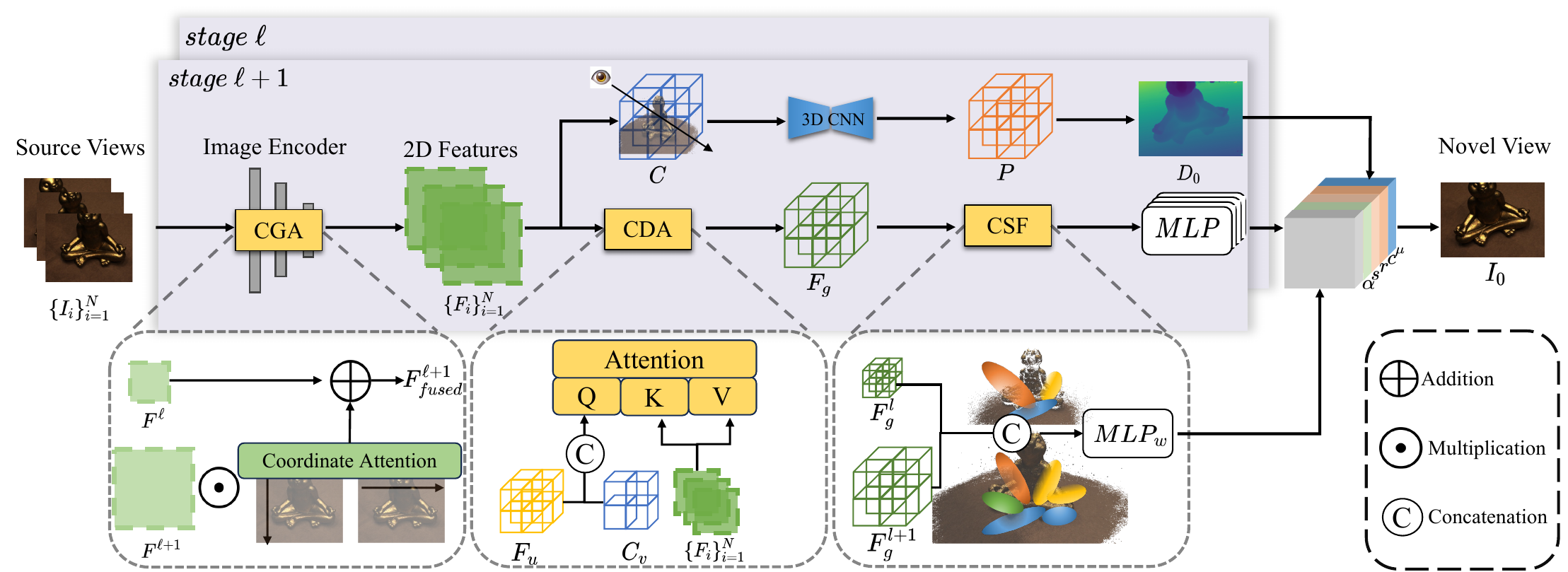}
    \vspace{5pt}
    \caption{\textbf{Overall architecture.} The proposed \ours is a coarse-to-fine framework that estimates depths and Gaussian representations from low resolution ($stage \ \ell$) to high resolution ($stage \ \ell+1$). It extracts features $\{\mathbf{F}_i\}_{i=1}^N$ from $N$ source images $\{\mathbf{I}_i\}_{i=1}^N$ using a feature pyramid network and CGA. These features are warped into the target camera frustum planes to construct the 3D cost volume $\mathbf{C}$. Regularization via 3D CNN yields the probability volume $\mathbf{P}$ that is regressed to generate the depth map $\mathbf{D}_{0}$, which serves as the Gaussian centers after unprojection. CDA is designed to enhance the interaction between 2D features from input views and 3D scene information from the cost volume $\mathbf{C}$ and the compressed features $\mathbf{F}_{u}$ from 2D features using a U-Net. The enhanced features $\mathbf{F}_{g}$ from CDA, are decoded to obtain the other Gaussian parameters (i.e., opacity, color, scale and rotation) via lightweight MLPs. By fusing the enhanced features from coarse and fine stages, the CSF produces weights for modifying Gaussian opacity. The output Gaussians are used to render the novel view $\mathbf{I}_{0}$.}
    \vspace{-10pt}
    \label{fig:pipeline}
\end{figure}

\section{Method}
\label{sec:method}

In this section, we reveal the details of our proposed framework. Our method is built upon MVSGaussian~\cite{mvsgaussian}, which estimates 3D Gaussian centers from multi-view stereo (MVS) and decodes Gaussian parameters via MLPs. We contribute three modules for image feature extraction, multi-view feature encoding, and multi-scale Gaussian fusion, enabling improved novel view synthesis without increasing complexity.

Specifically, given a set of posed source views $\{\mathbf{I}_{i}\}_{i=1}^N$, the goal of our framework is to render a target view $I_0$ from a novel viewpoint with predicted 3D Gaussians. As illustrated in Fig.~\ref{fig:pipeline}, we first extract multi-scale 2D features $\{\mathbf{F}_i\}_{i=1}^N$ from source images using a Feature Pyramid Network (FPN)~\cite{fpn}. These features are enhanced by our~\mdone(CGA) module and then warped to the target camera frustum to construct a cost volume $\mathbf{C}$ via differentiable homography. The cost volumes are passed through a 3D CNN to regularize the features and regress a depth map $\mathbf{D}_{0}$, which is unprojected to 3D space and used as the centers of pixel-aligned Gaussians. The enhanced features are further fed into our proposed~\mdtwo(CDA) module to learn cross-view and context-aware constraints, the output features $\mathbf{F}_{g}$ are then decoded to obtain the other Gaussian parameters (i.e., opacity, color, scale, and rotation) via lightweight MLPs. To bridge Gaussian representations across different scales, we propose~\mdthree(CSF) module to adaptively update Gaussian opacity $\alpha$ in finer stages, improving the rendering quality.

\vspace{-5pt}
\subsection{Image Feature Extraction with \mdonenospace}
\label{subsec:mdone}
MVSGaussian~\cite{mvsgaussian} employs standard FPN for feature extraction. Inspired by~\cite{zhang2023geomvsnet,hu2025icg}, we further enhance the expressiveness via introducing the \mdone(CGA) module, which models the relative importance of spatial positions between features, by applying two attention maps that capture long-range, feature-channel dependencies.
 
Specifically, given $N$ input images $\{\mathbf{I}_i\}_{i=1}^N$, we extract their multi-scale features $\{\mathbf{F}_{i}^{\ell}\}_{i=1}^N$ using a Feature Pyramid Network (FPN), where $\ell$ denotes the pyramid level. For clarity, since the same operations are performed independently for each image, we omit the image index $i$ in the following and denote the features at level $\ell$ by $\mathbf{F}^\ell$. At level $\ell$, the feature map $\mathbf{F}^{\ell}$ has lower resolution, while $\mathbf{F}^{\ell + 1}$ has double the height and width. We first compute the weights as $\mathbf{T}_{h} \in \mathbb{R}^{C \times H \times 1}$ and $\mathbf{T}_{w} \in \mathbb{R}^{C \times 1 \times W}$ according to Eq.~\ref{equ:attention}. Here, $H$ and $W$ represent the height and width of the features, respectively. Given a feature value in the $x$-th channel, the pooling operations along the height and width at position $(h, w)$ are defined as:
\begin{equation}
    \mathbf{T}_h(x, h, 1) = \frac{1}{W} \sum_{w=1}^{W} \mathbf{F}(x, h, w) \ ,
    \mathbf{T}_w(x, 1, w) = \frac{1}{H} \sum_{h=1}^{H} \mathbf{F}(x, h, w) \ .
    \label{equ:attention}
\end{equation}
Subsequently, the weights $\mathbf{T}_{h}$ and $\mathbf{T}_{w}$ are first reshaped and concatenated along the spatial dimension ($cat(\cdot, \cdot)$) to form $\mathbf{T}_\text{cat} \in \mathbb{R}^{C \times (H+W) \times 1}$. A convolution layer, followed by a sigmoid activation $\sigma(\cdot)$, is applied to merge the height and width information into a unified representation while normalizing the attention scores:
\begin{equation}
    \mathbf{T}_{\text{cat}} = cat(\mathbf{T}_h, \mathbf{T}_w) \in \mathbb{R}^{C \times (H + W) \times 1}
\end{equation}
\begin{equation}
    \mathbf{T}_{\text{attn}} = \sigma \left( \text{Conv1D}(\mathbf{T}_{\text{cat}}) \right) \in \mathbb{R}^{C \times (H + W) \times 1} \ .
\end{equation}
After fusion, the output $\mathbf{T}_{\text{attn}}$ is split to generate two distinct attention maps, $\mathbf{A}_{h} \in \mathbb{R}^{C \times H \times 1}$ and $\mathbf{A}_{w} \in \mathbb{R}^{C \times 1 \times W}$. Each attention map captures long-range dependencies along one spatial axis while maintaining coordinate cues along the orthogonal direction, enabling more efficient and interpretable representation learning.
The attention maps $\mathbf{A}_{h}$ and $\mathbf{A}_{w}$ are broadcasted to align with the shape of the input feature map $\mathbf{F}^{\ell+1}$ and applied multiplicatively to modulate the feature responses along the height and width dimensions:
\begin{equation}
    \mathbf{F}^{\ell + 1}_{A} = \mathbf{A}_{h} \odot \mathbf{A}_{w} \odot \mathbf{F}^{\ell + 1} \ .
    \label{equ:fusion}
\end{equation}
Meanwhile, the coarser feature map $\mathbf{F}^\ell$ from the upper pyramid level is upsampled and combined with the enhanced feature $\mathbf{F}^{\ell+1}_{A}$ via element-wise addition:
\begin{equation}
     \mathbf{F}^{\ell + 1}_{fused} = upsamp(\mathbf{F}^{\ell}) \oplus \mathbf{F}^{\ell + 1}_{A} \ ,
     \label{equ:branch}
\end{equation}
\noindent where $upsamp(\cdot)$, $\odot$, and $\oplus$ stand for the upsampling operation, element-wise multiplication (Hadamard product), and element-wise addition, respectively. By constructing 2D features with coordinate-aware long-range encoding, \mdone significantly improves global feature discrimination compared to standard convolutional operations, which are limited to local receptive fields.

\vspace{-5pt}
\subsection{Gaussian Parameter Prediction}
\paragraph{Gaussian Center Prediction from MVS}
After obtaining the coordinated-ware fused features $\mathbf{F} \in \mathbb{R}^{C \times H \times W}$ from source images, we follow a standard MVS pipeline~\cite{ibrnet,mvsgaussian}, to warp multi-scale features from source images into the target view by differentiable homographies, constructing a 3D cost volume $\mathbf{C}$ along fronto-parallel planes. The cost volume is processed by a 3D CNN to predict a depth probability volume $\mathbf{P}$, from which we regress the final depth map $\mathbf{D}_{0} \in \mathbb{R}^{H \times W \times 1}$, which are back-projected to 3D space with the target camera pose and served as the centers of new generated Gaussians, same as in MVSGaussian~\cite{mvsgaussian}. 

\paragraph{Feature Descriptor Construction with \mdtwonospace}
\label{subsec:mdtwo}
To encode multi-view features for a complete Gaussian representation, \cite{mvsgaussian} directly aggregates them through a simple pooling network, followed by a 2D U-Net for spatial enhancement. However, such encoding features mainly capture view-wise information and lack awareness of 3D context consistency, leading to discontinuities in feature representation. To address these limitations, we propose~\mdtwo (CDA) that jointly integrates 2D feature details and 3D volumetric consistency. 

In particular, we first extract per-pixel voxel features $\mathbf{C}_v \in \mathbb{R}^{HW \times 8}$ from the cost volume $\mathbf{C}$ through grid sampling, where $HW$ denotes the number of pixel locations after flattening the spatial dimensions. Simultaneously, we extract per-pixel image features across multiple source views, augmented with RGB values and ray direction differences, resulting in a feature tensor $\mathbf{F}_{img} \in \mathbb{R}^{HW \times N \times (C+3+4)}$, where $N$ denotes the number of source views, $C$ denotes the number of feature channels, the additional leading 3 channels correspond to RGB colors, and the trailing 4 channels encode ray direction differences.

To aggregate multi-view information at each pixel independently, we follow~\cite{mvsgaussian} to apply a lightweight U-Net $\Psi_u$ along the source view dimension, compressing $\mathbf{F}_{img}$ into a compact feature $\mathbf{F}_u \in \mathbb{R}^{HW \times 16}$:
\begin{equation}
    \mathbf{F}_u = \Psi_u(\mathbf{F}_{img})\,.
\end{equation}
We then concatenate the voxel consistency feature $\mathbf{C}_v$ and the aggregated image feature $\mathbf{F}_u$ to form a combined representation:
\begin{equation}
    \mathbf{F}_c = cat(\mathbf{F}_u, \mathbf{C}_v)\,,
\end{equation}
where $\mathbf{F}_c \in \mathbb{R}^{HW \times 24}$. This fused feature integrates both fine-grained 2D appearance information and multi-view geometric consistency, enhancing the robustness of 3D representation.

Finally, we leverage $\mathbf{F}_c$ as the query input to an attention-based aggregation module $\Psi_t$, which retrieves and integrates 2D features across multiple views:
\begin{equation}
    \mathbf{F}_g = \Psi_t(\mathbf{F}_c, \{\mathbf{F}_i\}_{i=1}^N)\,,
\end{equation}
where $\mathbf{F}_c$ serves as the query, and $\{\mathbf{F}_i\}_{i=1}^N$ provide the key and value sequences. This cross-view attention mechanism enables the encoded features to capture both spatially consistent 3D geometry and rich appearance details, leading to improved spatial-aware reconstruction.

\vspace{-5pt}
\paragraph{Gaussian Attribute Prediction with \mdthreenospace}
\label{subsec:mdthree}
The encoded features $\mathbf{F}_{g}$ are used to predict the parameters of each Gaussian. Specifically, each Gaussian is defined by a set of attributes, $\{\mu, s, r, \alpha, c\}$, where $\mu$ denotes the center, $s$ the scale, $r$ the rotation, $\alpha$ the opacity, and $c$ the color. As mentioned before, the center $\mu$ is determined by back-projecting pixel locations according to the regressed depth map $\mathbf{D}_{0}$. The remaining parameters are decoded from $\mathbf{F}_g$ as:
\begin{equation}
\label{eq:scale}
\begin{aligned}
\left\{
\begin{aligned}
s &= Softplus(MLP_s(\mathbf{F}_g))\, \\
r &= Norm(MLP_r(\mathbf{F}_g))\, \\
\alpha &= Sigmoid(MLP_{\alpha}(\mathbf{F}_g))\, \\
c &= Sigmoid(MLP_c(\mathbf{F}_g))\,
\end{aligned}
\right.,
\end{aligned}
\end{equation}
where $MLP_s$, $MLP_r$, $MLP_\alpha$, and $MLP_c$ are dedicated heads for estimating scale, rotation, opacity, and color, respectively.

Previous works~\cite{mvpgs,mvsgaussian} adopt cascaded multi-view stereo designs to predict depth maps from coarse to fine. However, they produce fixed-resolution images with single-scale features, without explicitly modeling the relationships across different scales of Gaussians. To address this, we propose the \mdthree (CSF) module, which modulates the opacity of Gaussians across scales to enhance consistency.

Specifically, given the Gaussian features from two adjacent stages, $\mathbf{F}_{g}^{\ell}$ and $\mathbf{F}_{g}^{\ell+1}$, we first upsample $\mathbf{F}_{g}^{\ell}$ to match the spatial resolution of $\mathbf{F}_{g}^{\ell+1}$, and then concatenate them along the feature dimension:
\begin{equation}
    \mathbf{F}^{\ell+1}_{g} = cat(upsamp(\mathbf{F}_{g}^{\ell}), \mathbf{F}_{g}^{\ell+1})\,,
\end{equation}
where $upsamp(\cdot)$ denotes spatial interpolation, and $cat(\cdot, \cdot)$ denotes feature concatenation. The concatenated feature is passed through a lightweight MLP, $MLP_w$, to produce a modulation weight $\mathbf{w}^{\ell+1}$:
\begin{equation}
    \mathbf{w}^{\ell+1} = MLP_w(\mathbf{F}^{\ell+1}_{g})\,,
\end{equation}
which is used to adjust the opacity by element-wise multiplication:
\begin{equation}
    \alpha = \alpha \odot \mathbf{w}^{\ell+1}\,.
\end{equation}
With the updated opacity $\alpha$ and the other Gaussian attributes $\{\mu, s, r, c\}$, novel views $\mathbf{I}_{0}$ can be rendered via rasterization. Although in principle all Gaussian parameters could be refined across scales, we empirically find that updating only the opacity achieves a good balance between performance and computational efficiency.

\vspace{-5pt}
\subsection{Loss Function}
\label{subsec:loss function}
Following~\cite{mvsgaussian}, our model is trained using images as the only source of supervision. We combine a standard pixel-wise error (MSE) $\mathcal{L}_{\mathrm{pixel}}$ with additional perceptual~\cite{lpips} (VGG16) $\mathcal{L}_{\mathrm{feature}}$ and structural similarity~\cite{ssim} (SSIM) $\mathcal{L}_{\mathrm{structure}}$ losses to improve generalization. 
Our framework consists of two stages, corresponding to a coarse and a fine reconstruction level. The overall loss is computed as the weighted sum of the losses at each stage:
\begin{equation}
\label{eq:overall_loss}
\mathcal{L}_{\mathrm{total}} = \sum_{\ell} \gamma^{(\ell)} \left( \mathcal{L}_{\mathrm{pixel}} + \beta_s \cdot \mathcal{L}_{\mathrm{structure}} + \beta_p \cdot \mathcal{L}_{\mathrm{feature}} \right) \,,
\end{equation}
where $\mathcal{L}_{\mathrm{total}}$ represents the cumulative loss, and $\gamma^{(\ell)}$ denotes the importance weight of the $\ell$-th stage loss. The loss at each stage consists of three components: a pixel-wise term $\mathcal{L}_{\mathrm{pixel}}$, a structure-based term $\mathcal{L}_{\mathrm{structure}}$ scaled by $\beta_s$, and a feature-based term $\mathcal{L}_{\mathrm{feature}}$ scaled by $\beta_p$.

\begin{figure}[htb]
  \centering
  \includegraphics[width=0.96\linewidth]{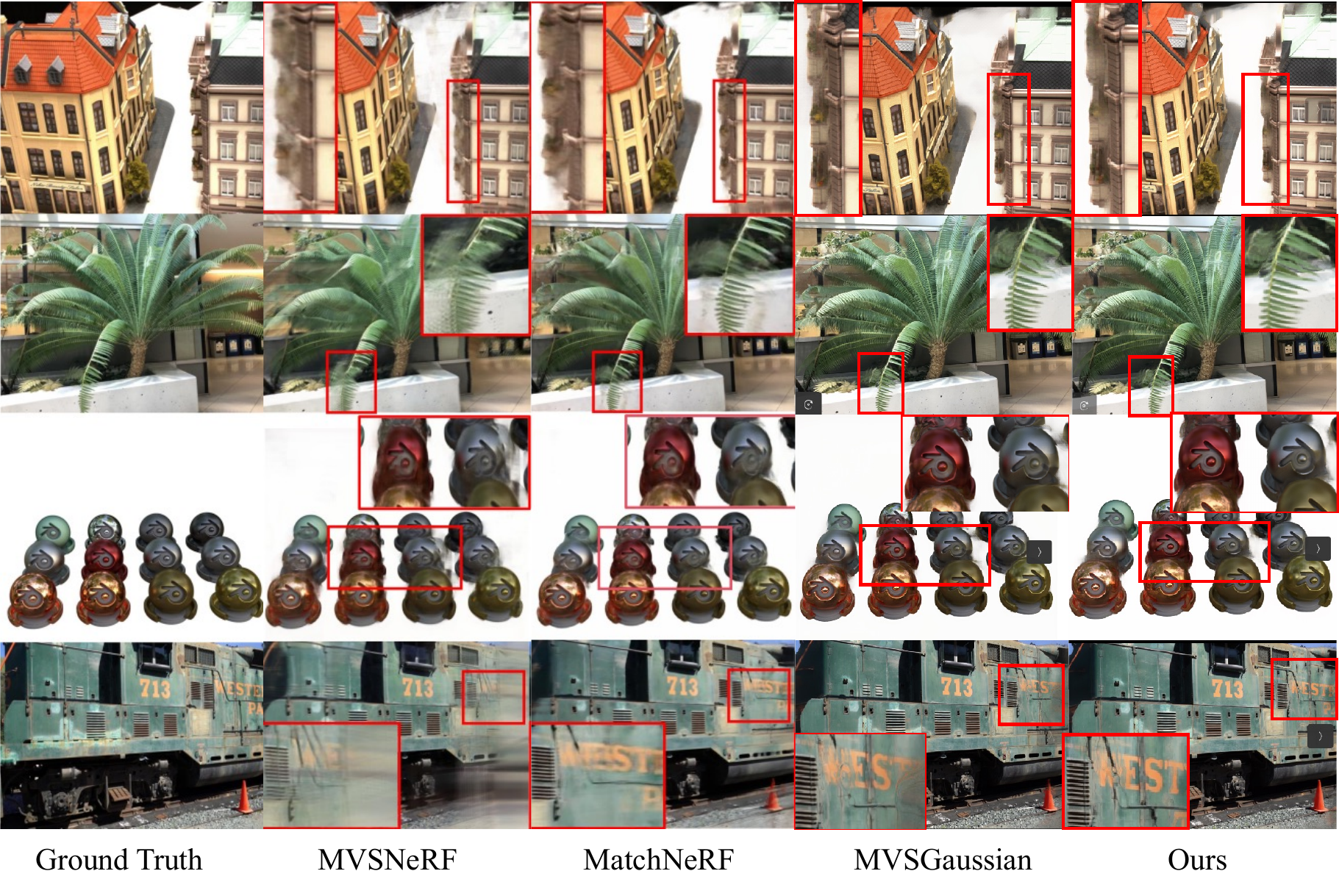}
  \vspace{-5pt}
  \caption{\textbf{Qualitative comparison with state-of-the-art methods~\cite{mvsnerf,matchnerf,mvsgaussian} using 3 input views on DTU~\cite{dtu}, Real Forward-facing~\cite{llff}, NeRF Synthetic~\cite{nerf}, and Tanks and Temples~\cite{tanks}, arranged top to bottom.}}
  \vspace{-10pt}
  \label{fig:vis_gen}
\end{figure}

%% file: sec/4_experiments.tex
\section{Experiments}
\label{sec:experiments}

\vspace{-5pt}
\subsection{Experimental Setup}
\label{subsec:experimental setup}
\noindent \textbf{Datasets.}
As in~\cite{mvsnerf,mvsgaussian}, we train our model using the DTU training set~\cite{dtu} and evaluate on the DTU test set, applying the same dataset split configuration used in~\cite{mvsgaussian}. We evaluate our model directly on the Real Forward-facing~\cite{llff}, NeRF Synthetic~\cite{nerf}, and Tanks and Temples~\cite{tanks} datasets, without further training. For each scene, we select the same views as target views as established in~\cite{mvsgaussian}, and the remaining images may serve as source views based on their distance from the target view. The quality of synthesized views is measured by broadly adopted PSNR, SSIM~\cite{ssim}, and LPIPS~\cite{lpips} metrics.

\begin{wraptable}[10]{r}{0.48\linewidth}
\centering
\setlength{\tabcolsep}{2pt}
\resizebox{1\linewidth}{!}{
\begin{tabular}{@{}lccccccc@{}}
\toprule
\multirow{2}{*}{Method} & \multicolumn{3}{c}{2-view} & \multicolumn{3}{c}{3-view} \\ 
\cmidrule(lr){2-4} \cmidrule(lr){5-7}
 & PSNR $\uparrow$ & SSIM $\uparrow$ & LPIPS $\downarrow$ & PSNR $\uparrow$ & SSIM $\uparrow$ & LPIPS $\downarrow$ \\
\midrule
MVSNeRF~\cite{mvsnerf} & \cellcolor{softyellow}{24.03} & 0.914 & 0.192 & 26.63 & 0.931 & 0.168 \\
MatchNeRF~\cite{matchnerf} & \cellcolor{softred}{25.03} & \cellcolor{softorange}{0.919} & \cellcolor{softyellow}{0.181} & \cellcolor{softyellow}{26.91} & \cellcolor{softyellow}{0.934} & 0.159 \\
PixelSplat~\cite{pixelsplat} & 14.01 & 0.662 & 0.389 & - & - & - \\
MVSplat~\cite{mvsplat} & 13.94 & 0.473 & 0.385 & 14.29 & 0.508 & 0.370 \\
MVPGS~\cite{mvpgs} & 17.55 & 0.823 & 0.147 & 20.65 & 0.877 & \cellcolor{softyellow}{0.099} \\
MVSGaussian~\cite{mvsgaussian} & 23.83 & \cellcolor{softyellow}{0.918} & \cellcolor{softorange}{0.121} & \cellcolor{softorange}{27.03} & \cellcolor{softorange}{0.959} & \cellcolor{softorange}{0.084} \\
Ours & \cellcolor{softorange}{24.27} & \cellcolor{softred}{0.936} & \cellcolor{softred}{0.113} & \cellcolor{softred}{27.87} & \cellcolor{softred}{0.962} & \cellcolor{softred}{0.077} \\
\bottomrule
\end{tabular}}
\vspace{-6pt}
\caption{\textbf{Quantitative results of generalization on DTU~\cite{dtu}.}}
\label{Tab:gen_dtu}
\end{wraptable}

\noindent \textbf{Implementation Details}
\label{subsec:implementation details}
Our framework is built on MVSGaussian~\cite{mvsgaussian}, which employs a two-stage cascaded framework. Following~\cite{mvsgaussian}, we train the model with 3 source views and additionally test its performance with a varying number of input views (2, 3, 4). 
For depth estimation, we sample $64$ and $8$ depth hypotheses for the coarse and fine stages, respectively. 
For loss function, we set $\beta_s=0.1$, $\beta_p=0.05$, $\gamma^{(1)}=0.5$ and $\gamma^{(2)}=1$ in Eq.~\ref{eq:overall_loss}.
The model is trained using the Adam optimizer~\cite{adam} on a single RTX 4090 GPU.

\noindent \textbf{Baselines.}
We compare our framework with state-of-the-art generalizable Gaussian methods~\cite{pixelsplat,mvsplat,mvpgs,mvsgaussian}, as well as generalizable NeRF methods~\cite{mvsnerf,matchnerf}. We use their reported results if available. However, due to the difference between pipelines and experimental settings (\ie views, datasets, and resolutions), some results are not available. In such cases, we generate the results using their released code and pre-trained model on other datasets. Note that we also re-train our baseline model (MVSGaussian~\cite{mvsgaussian}) from scratch on our machine to eliminate environmental differences for fair comparison. For all tables, we use color to indicate rankings: \colorbox{softred}{red} for first place, \colorbox{softorange}{orange} for second, and \colorbox{softyellow}{yellow} for third.

\begin{table}[htb]
\centering
\setlength{\tabcolsep}{5pt}
\resizebox{1\linewidth}{!}{
\begin{tabular}{@{}lccccccccccc@{}}
\toprule
\multirow{2}{*}{Method} & \multirow{2}{*}{Views} & \multicolumn{3}{c}{Real Forward-facing~\cite{llff}} & \multicolumn{3}{c}{NeRF Synthetic~\cite{nerf}} & \multicolumn{3}{c}{Tanks and Temples~\cite{tanks}}\\ 
\cmidrule(lr){3-5}\cmidrule(lr){6-8}\cmidrule(lr){9-11}
 & & PSNR $\uparrow$ & SSIM $\uparrow$ & LPIPS $\downarrow$ & PSNR $\uparrow$ & SSIM $\uparrow$ & LPIPS $\downarrow$ & PSNR $\uparrow$ & SSIM $\uparrow$ & LPIPS $\downarrow$ \\
\midrule
MVSNeRF~\cite{mvsnerf} & \multirow{6}{*}{2} & 20.22 & 0.763 & 0.287 & 20.56 & 0.856 & 0.243 & 18.92 & 0.756 & 0.326 \\
MatchNeRF~\cite{matchnerf} & & \cellcolor{softyellow}{20.59} & \cellcolor{softyellow}{0.775} & \cellcolor{softyellow}{0.276} & \cellcolor{softyellow}{20.57} & \cellcolor{softyellow}{0.864} & \cellcolor{softyellow}{0.200} & \cellcolor{softyellow}{19.88} & \cellcolor{softyellow}{0.773} & \cellcolor{softyellow}{0.334} \\
MVSplat~\cite{mvsplat}  & & 15.32 & 0.370 & 0.422 & 12.59 & 0.665 & 0.282 & 16.35 & 0.490 & 0.355 \\
MVPGS~\cite{mvpgs}  & & 18.53 & 0.607 & 0.280 & - & - & - & 16.33 & 0.687 & 0.373 \\
MVSGaussian~\cite{mvsgaussian} & & \cellcolor{softred}{22.58} & \cellcolor{softorange}{0.817} & \cellcolor{softred}{0.192} & \cellcolor{softorange}{23.56} & \cellcolor{softorange}{0.929} & \cellcolor{softorange}{0.122} & \cellcolor{softorange}{21.05} & \cellcolor{softorange}{0.830} & \cellcolor{softorange}{0.222} \\
Ours & & \cellcolor{softorange}{22.32} & \cellcolor{softred}{0.818} & \cellcolor{softorange}{0.201} & \cellcolor{softred}{23.83} & \cellcolor{softred}{0.930} & \cellcolor{softred}{0.109} & \cellcolor{softred}{21.58} & \cellcolor{softred}{0.832} & \cellcolor{softred}{0.221} \\
\midrule 
MVSNeRF~\cite{mvsnerf}  &\multirow{5}{*}{3} & 21.93 & 0.795 & 0.252 & \cellcolor{softyellow}{23.62} & \cellcolor{softyellow}{0.897} & 0.176 & \cellcolor{softyellow}{20.87} & \cellcolor{softyellow}{0.823} & \cellcolor{softyellow}{0.260} \\
MatchNeRF~\cite{matchnerf} & & \cellcolor{softyellow}{22.43} & \cellcolor{softyellow}{0.805} & 0.244 & 23.20 & \cellcolor{softyellow}{0.897} & \cellcolor{softyellow}{0.164} & 20.80 & 0.793 & 0.300 \\
MVPGS~\cite{mvpgs}  & & 19.91 & 0.696 & \cellcolor{softyellow}{0.229} & - & - & - & 18.60 & 0.607 & 0.467 \\
MVSGaussian~\cite{mvsgaussian} & & \cellcolor{softorange}{23.87} & \cellcolor{softorange}{0.856} & \cellcolor{softorange}{0.169} & \cellcolor{softorange}{25.31} & \cellcolor{softorange}{0.943} & \cellcolor{softorange}{0.098} & \cellcolor{softorange}{22.70} & \cellcolor{softorange}{0.872} & \cellcolor{softorange}{0.187} \\
Ours & & \cellcolor{softred}{23.97} & \cellcolor{softred}{0.857} & \cellcolor{softred}{0.167} & \cellcolor{softred}{26.14} & \cellcolor{softred}{0.946} & \cellcolor{softred}{0.079} & \cellcolor{softred}{23.30} & \cellcolor{softred}{0.879} & \cellcolor{softred}{0.178} \\
\midrule
MVPGS~\cite{mvpgs}  &\multirow{3}{*}{4} & \cellcolor{softyellow}{21.28} & \cellcolor{softyellow}{0.750} & \cellcolor{softyellow}{0.180} & - & - & - & \cellcolor{softyellow}{20.27} & \cellcolor{softyellow}{0.679} & \cellcolor{softyellow}{0.371} \\
MVSGaussian~\cite{mvsgaussian} & & \cellcolor{softorange}{24.01} & \cellcolor{softorange}{0.864} & \cellcolor{softorange}{0.175} & \cellcolor{softorange}{24.81} & \cellcolor{softorange}{0.938} & \cellcolor{softorange}{0.106} & \cellcolor{softorange}{21.85} & \cellcolor{softorange}{0.860} & \cellcolor{softorange}{0.209} \\
Ours & & \cellcolor{softred}{24.16} & \cellcolor{softred}{0.867} & \cellcolor{softred}{0.174} & \cellcolor{softred}{25.29} & \cellcolor{softred}{0.941} & \cellcolor{softred}{0.087} & \cellcolor{softred}{22.35} & \cellcolor{softred}{0.866} & \cellcolor{softred}{0.204} \\
\bottomrule
\end{tabular}}
\vspace{-5pt}
\caption{\textbf{Quantitative results of generalization on Real Forward-facing~\cite{llff}, NeRF Synthetic~\cite{nerf}, and Tanks and Temples~\cite{tanks} datasets.}}
\vspace{-10pt}
\label{Tab:gen_llff}
\end{table}

\begin{table}[htb]
\centering
\setlength{\tabcolsep}{5pt}
\resizebox{1\linewidth}{!}{
\begin{tabular}{@{}lccccccccccc@{}}
\toprule
\multirow{2}{*}{Method} & \multirow{2}{*}{Views} & \multicolumn{3}{c}{Real Forward-facing~\cite{llff}} & \multicolumn{3}{c}{NeRF Synthetic~\cite{nerf}} & \multicolumn{3}{c}{Tanks and Temples~\cite{tanks}}\\ 
\cmidrule(lr){3-5}\cmidrule(lr){6-8}\cmidrule(lr){9-11}
 & & PSNR $\uparrow$ & SSIM $\uparrow$ & LPIPS $\downarrow$ & PSNR $\uparrow$ & SSIM $\uparrow$ & LPIPS $\downarrow$ & PSNR $\uparrow$ & SSIM $\uparrow$ & LPIPS $\downarrow$ \\
\midrule
PixelSplat~\cite{pixelsplat} & \multirow{3}{*}{2} & \cellcolor{softred}{22.99} & \cellcolor{softyellow}{0.810} & \cellcolor{softyellow}{0.190} & \cellcolor{softyellow}{15.77} & \cellcolor{softyellow}{0.755} & \cellcolor{softyellow}{0.314} & \cellcolor{softyellow}{19.40} & \cellcolor{softyellow}{0.689} & \cellcolor{softyellow}{0.223} \\
MVSGaussian~\cite{mvsgaussian} & & \cellcolor{softorange}{22.87} & \cellcolor{softorange}{0.823} & \cellcolor{softorange}{0.159} & \cellcolor{softorange}{24.00} & \cellcolor{softorange}{0.928} & \cellcolor{softorange}{0.121} & \cellcolor{softorange}{21.81} & \cellcolor{softorange}{0.842} & \cellcolor{softorange}{0.176} \\
Ours & & \cellcolor{softorange}{22.87} & \cellcolor{softred}{0.825} & \cellcolor{softred}{0.156} & \cellcolor{softred}{24.51} & \cellcolor{softred}{0.929} & \cellcolor{softred}{0.098} & \cellcolor{softred}{22.34} & \cellcolor{softred}{0.847} & \cellcolor{softred}{0.172} \\
\bottomrule
\end{tabular}}
\vspace{-5pt}
\caption{\textbf{Comparison with PixelSplat~\cite{pixelsplat}.} Given the high memory requirements of PixelSplat~\cite{pixelsplat}, the evaluations are conducted on low-resolution ($512\times512$) images.}
\vspace{-10pt}
\label{Tab:gen_llff_pixelsplat}
\end{table}

\vspace{-5pt}
\subsection{Generalization Results}
\label{subsec: generalization results}
We first demonstrate the results on the DTU~\cite{dtu} test set. As shown in Fig.~\ref{fig:vis_gen}, our results preserve more scene details with fewer artifacts. Quantitative results in Table~\ref{Tab:gen_dtu} show that we achieve the best scores in all settings except for 2-view PSNR. This is mainly because MatchNeRF~\cite{matchnerf} adopts a pre-trained model~\cite{xu2022gmflow} that is specially designed to capture accurate optical flow between two input images as a match prior.

To further assess the generalization ability of our method, we conduct experiments on Real Forward-facing~\cite{llff}, NeRF Synthetic~\cite{nerf}, and Tanks and Temples~\cite{tanks}, with results shown in Fig.~\ref{fig:vis_gen} and Table~\ref{Tab:gen_llff}. Additional visualizations are provided in the supplementary materials. 
Note that, due to their fixed architectures, we exclude 4-view results for MVSNeRF~\cite{mvsnerf} and MatchNeRF~\cite{matchnerf} and provide only 2-view results for MVSplat~\cite{mvsplat}. For PixelSplat~\cite{pixelsplat} with large memory consumption, we compare our method with it in lower-resolution images separately in Table~\ref{Tab:gen_llff_pixelsplat}.
As MVPGS~\cite{mvpgs} requires per-scene training assisted by monocular depth and masks, we re-train it a fixed number of times for fair comparison.

As demonstrated in Table~\ref{Tab:gen_llff}, our method achieves top-ranking performance with 3 and 4 input views. A slight performance drop at 4 views is observed for both our method and MVSGaussian~\cite{mvsgaussian}, likely due to limited view overlap, as discussed in prior MVS studies~\cite{unimvs,zhang2023geomvsnet}.
Overall, our approach generalizes well without any additional training, benefiting from enhanced feature extraction across 2D and 3D spaces.

\vspace{-5pt}
\subsection{Depth Estimation Results}
\label{subsec:depth reconstruction results}
\begin{wrapfigure}{r}{0.45\textwidth}
    \centering
    \includegraphics[width=0.45\textwidth]{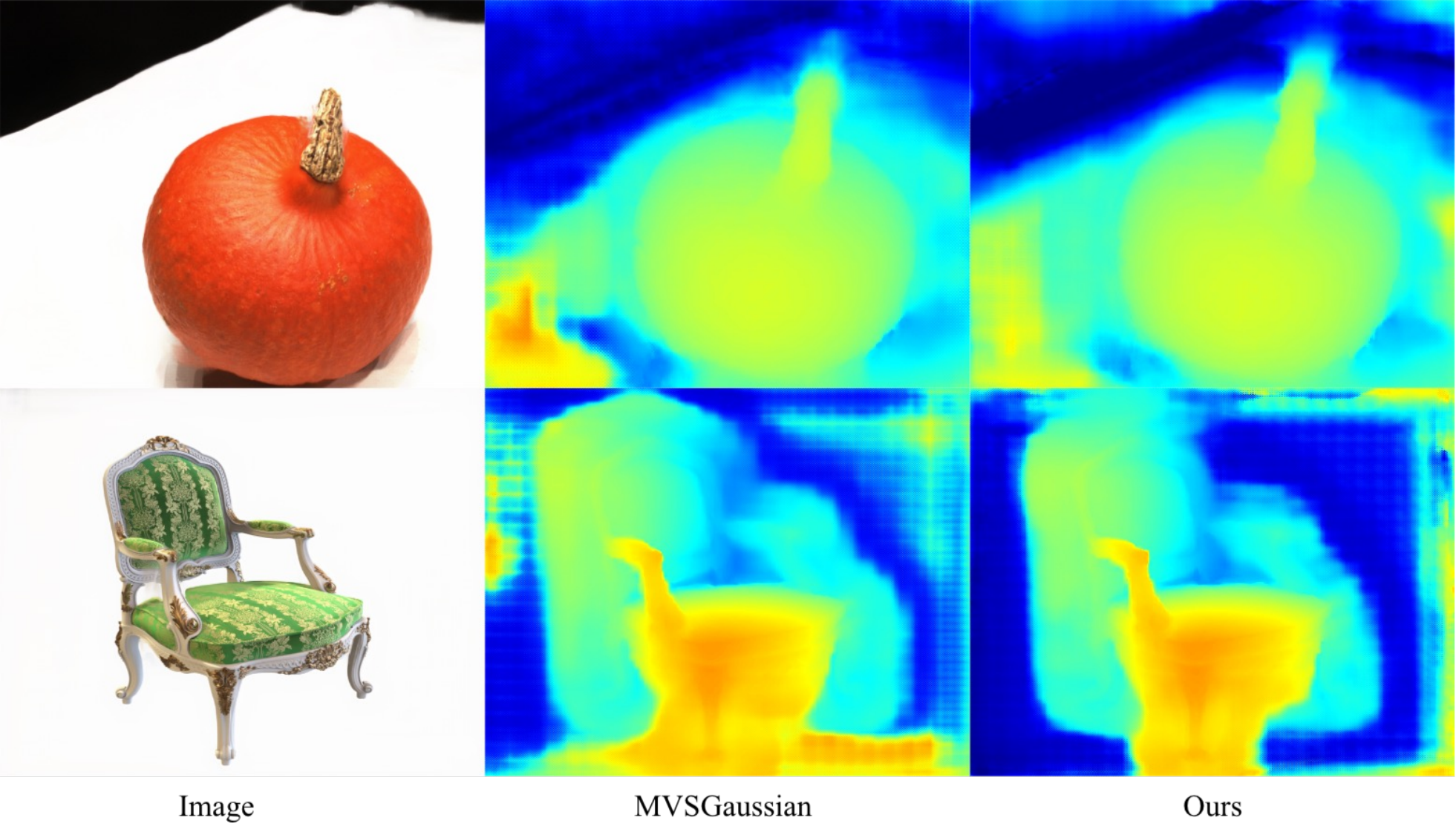}
    \vspace{-5pt}
    \caption{\textbf{Qualitative comparison of depth maps with MVSGaussian~\cite{mvsgaussian}.}}
    \vspace{-10pt}
    \label{fig:depth_vis}
\end{wrapfigure}

Following~\cite{mvsnerf,mvsgaussian}, we also evaluate depth map estimation. As illustrated in Fig.~\ref{fig:depth_vis}, our method produces sharper and more accurate depth boundaries, despite being trained without ground-truth supervision. Quantitative comparisons in Table~\ref{Tab:depth} further confirm our superiority. 
An accurate depth map, as an intermediate output in our model, indicates that our method generates more precise encoded features for Gaussian parameters decoder. These refined features form a solid foundation for the subsequent estimation of Gaussian parameters, enhancing the reliability and quality of the rendered scene.

\vspace{-5pt}
\subsection{Ablation Study}
\label{subsec:ablation study}
As shown in Table~\ref{Tab:ablation_module}, we perform ablation studies on the DTU~\cite{dtu} test set (3-view input) to evaluate the effectiveness of each module. Using only \mdone (CGA) yields limited improvements (No.1), while \mdtwo (CDA) significantly boosts PSNR (No.2). Combining CGA and CDA (No.4) further improves SSIM and LPIPS, indicating complementary effects. \mdthree (CSF) alone also enhances results (No.3), and combining it with CGA (No.5) or CDA (No.6) leads to consistent gains. Finally, integrating all three modules (No.7) achieves the best overall performance. Corresponding qualitative comparisons are available in the supplementary material.

\begin{table}[htb]
\centering
\begin{minipage}{0.48\textwidth}
\setlength{\tabcolsep}{5pt}
\resizebox{1\linewidth}{!}{
\begin{tabular}{@{}lcccccc@{}}
\toprule
\multirow{2}{*}{Method} & \multicolumn{3}{c}{Novel view} \\ 
\cmidrule(lr){2-4}\cmidrule(lr){5-7} & Abs err $\downarrow$ & Acc(2)$\uparrow$ & Acc(10)$\uparrow$ \\
\midrule
PixelNeRF~\cite{pixelnerf} & 47.8 & 0.039 & 0.187 \\
IBRNet~\cite{ibrnet} & 324 & 0.000 & 0.866  \\
MVSNeRF~\cite{mvsnerf} & \cellcolor{softyellow}{7.00} & \cellcolor{softyellow}{0.717} & \cellcolor{softyellow}{0.866} \\
MVSGaussian~\cite{mvsgaussian} & \cellcolor{softorange}{4.06} & \cellcolor{softorange}{0.820} & \cellcolor{softorange}{0.936}  \\
Ours & \cellcolor{softred}{3.79} & \cellcolor{softred}{0.832} & \cellcolor{softred}{0.943}
\\
\bottomrule
\end{tabular}}
\vspace{-5pt}
\caption{\textbf{Quantitative results of depth estimation on DTU~\cite{dtu}.} ``Abs err" denotes the mean absolute error (mm), while ``Acc(X)" indicates the percentage of pixels with an error below X mm.}
\vspace{-10pt}
\label{Tab:depth}
\end{minipage}
\begin{minipage}{0.48\textwidth}
\centering
\setlength{\tabcolsep}{5pt}
\resizebox{1\linewidth}{!}{
\begin{tabular}{@{}lccc|ccc@{}}
\toprule
& CGA & CDA & CSF & PSNR$\uparrow$ & SSIM$\uparrow$ & LPIPS$\downarrow$ \\
\midrule
baseline & - & - & - & 27.03 & 0.959 & 0.084 \\
No.1 & \checkmark & - & - & 27.07 & 0.959 & 0.084 \\
No.2 & - & \checkmark & - & 27.34 & 0.959 & 0.084 \\
No.3 & - & - & \checkmark & 27.31 & 0.960 & 0.080 \\
No.4 & \checkmark & \checkmark & - & 27.42 & 0.960 & 0.078 \\
No.5 & \checkmark & - & \checkmark & 27.58 & 0.960 & 0.079 \\
No.6 & - & \checkmark & \checkmark & 27.79 & 0.961 & 0.077 \\
No.7 & \checkmark & \checkmark & \checkmark & \textbf{27.87} & \textbf{0.962} & \textbf{0.077} \\
\bottomrule
\end{tabular}}
\vspace{-5pt}
\caption{\textbf{Ablation studies on combination of modules.} The PSNR, SSIM and LPIPS are the metrics on DTU datasets~\cite{dtu}. The baseline method is MVSGaussian~\cite{mvsgaussian}.}
\vspace{-10pt}
\label{Tab:ablation_module}
\end{minipage}
\end{table}

%% file: sec/5_conclusion.tex
\section{Conclusion}
\label{sec:conclusion}
In this work, we propose \oursnospace, a generalizable Gaussian Splatting framework that enhances feature encoding with context-aware, cross-dimensional, and cross-scale information, without relying on additional supervisory signals. By leveraging lightweight feature aggregation and flexible multi-view constraints, \ours achieves consistent improvements in cross-dataset generalization over existing state-of-the-art methods. Extensive experiments validate the effectiveness of our approach. 
%
%
However, several important challenges remain unaddressed. First, our current framework does not explicitly handle challenging real-world conditions such as domain shifts across heterogeneous camera systems or degraded inputs caused by motion blur and image noise. Second, our method, like other 3D cost volume–based approaches, may face limitations in wide-baseline scenarios and under view extrapolation, where target views lie outside the range of input observations. Addressing these scenarios can be explored in future work.

%% file: sec/X_suppl.tex
\clearpage
\setcounter{page}{1}

In the supplementary material, we present more details that are not included in the main text, including:

\begin{itemize}
    \item Extended ablation studies, such as qualitative comparison and varying numbers of training views.
    \item Resource consumption analysis, examining module-wise time overhead in our framework and comparing runtime and memory usage with other methods.
    \item Results after per-scene optimization of Gaussians.
    \item Additional visualizations, including rendered images, depth maps, and source view selection examples.
\end{itemize}

\section{More Ablations}
\subsection{Qualitative visualizations for ablations}
\begin{figure}[htb]
    \centering
    \includegraphics[width=0.98\linewidth]{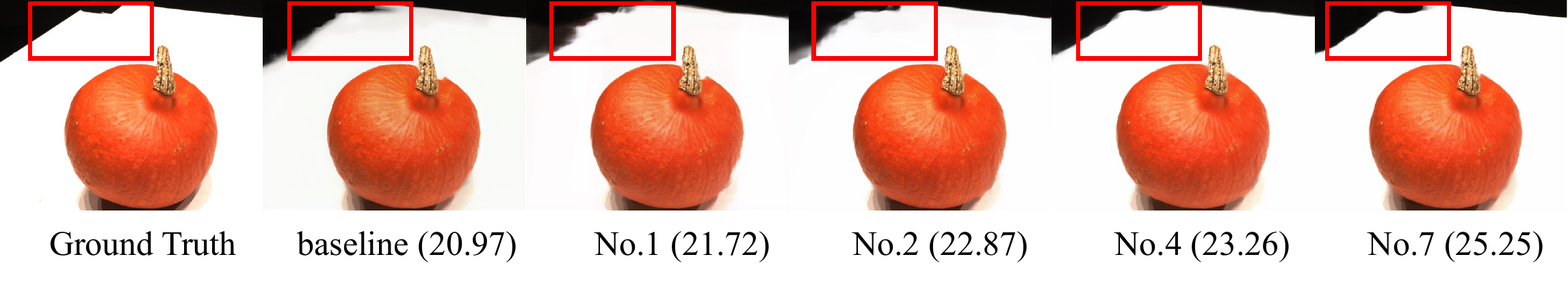}
    \caption{\textbf{Visualization of ablation study.} The numbers in the figures refer to the ablation numbers in the manuscript, with the corresponding PSNR values displayed for each image.}
    \label{fig:vis_ablation}
\end{figure}

Fig.~\ref{fig:vis_ablation} shows rendered images from the ablation in Table~\ref{Tab:ablation_module} to demonstrate the role of the module, mainly for the effectiveness of CGA. The baseline renders the foreground well but has blurred edges and incorrect table colors. CGA alone (No.1) does not improve much visually, as it is tightly coupled with subsequent modules like CDA to enhance cross-dimensional interactions.
Using CDA alone (No.2) improves color rendering and edge geometry but still results in blurred edges and some artifacts. When combining both modules (No.4), these issues are reduced, with sharper edges and preserved geometry. Integrating the CSF further enhances background sharpness (No.7). 

\subsection{Number of training views}
In the main paper, we trained our model with 3 input images as source views and evaluated it using varying numbers of source views. Here, we perform an ablation experiment by training the model with different numbers of input views. In Table~\ref{Tab:ablation_view}, both training and testing are conducted with the same number of inputs—for instance, training with 2 source views and testing with 2 source views. 

As shown in Table~\ref{Tab:ablation_view}, across all configurations, our method achieves strong performance, as evidenced by high PSNR, SSIM scores, and low LPIPS values. Results are generally better when the number of training views matches the testing views (marked with $*$), suggesting that training-specific configurations benefit from consistency. Taking training views into consideration, now we have the best PSNR performance in Table~\ref{Tab:gen_dtu} under the 2-views setting.

\begin{table}[htb]
\centering
\setlength{\tabcolsep}{5pt}
\resizebox{0.6\linewidth}{!}{
\begin{tabular}{@{}lccccc@{}}
\toprule
\multirow{2}{*}{Method} & \multirow{2}{*}{Views} & \multicolumn{3}{c}{DTU~\cite{dtu}} \\ 
\cmidrule(lr){3-5}
 & & PSNR $\uparrow$ & SSIM $\uparrow$ & LPIPS $\downarrow$ \\
\midrule
Ours & \multirow{2}{*}{2}  & 24.27 & 0.936 & 0.113 \\
Ours$^{*}$ & &  25.14 & 0.939 & 0.102 \\
\midrule 
Ours$^{*}$ & \multirow{1}{*}{3} & 27.87 & 0.962 & 0.077 \\
\midrule
Ours & \multirow{2}{*}{4}  & 27.63 & 0.962 & 0.081 \\
Ours$^{*}$ & & 27.95 & 0.962  & 0.079 \\
\bottomrule
\end{tabular}}
\vspace{5pt}
\caption{\textbf{Ablation studies on the number of training views on DTU~\cite{dtu}.} ``Ours$^{*}$" means training views match the testing views.}
\label{Tab:ablation_view}
\end{table}

\begin{table}[t]
\centering
\setlength{\tabcolsep}{5pt}
\resizebox{0.6\linewidth}{!}{
\begin{tabular}{@{}lccc@{}}
\toprule
\multirow{1}{*}{Method} & & \multirow{1}{*}{Mem (GB)$\downarrow$}  & \multirow{1}{*}{FPS$\uparrow$} \\
\midrule
MVSNeRF~\cite{mvsnerf}  &  & - & 0.42 \\
MatchNeRF~\cite{matchnerf} & & - & 1.04\\
PixelSplat~\cite{pixelsplat} & & 11.83$^{*}$ & 1.13$^{*}$ \\
MVSplat~\cite{mvsplat} & & 1.76 & 17.80 \\
MVPGS~\cite{mvpgs} & & 0.28 & 21.93 \\
MVSGaussian~\cite{mvsgaussian} & & 0.88/0.87$^{*}$ & 21.50/24.50$^{*}$ \\
Ours & & 1.01/0.89$^{*}$ & 14.57/15.50$^{*}$ \\
\bottomrule
\end{tabular}}
\vspace{5pt}
\caption{\textbf{Resources consumption on DTU~\cite{dtu}.} FPS and Mem are measured under a 3-view input, while FPS$^*$ and Mem$^*$ are measured under a 2-view input.}
\label{Tab:runtime_comparison}
\end{table}

\begin{table}
\centering
\resizebox{0.6\linewidth}{!}{
\begin{tabular}{@{}cccc@{}}
\toprule
\multicolumn{2}{c}{Modules} & coarse stage & fine stage \\
\midrule
\multicolumn{2}{c}{Image Encoder} & \multicolumn{2}{c}{1.25} \\
\midrule
\multicolumn{2}{c}{Depth Estimation} & 9.02 & 2.65 \\
\midrule
\multicolumn{2}{c}{CDA} & 0.83 & 1.16 \\
\midrule
\multicolumn{2}{c}{CSF} & - & 0.31 \\
\midrule
\multicolumn{2}{c}{Gaussian Rendering} & - & 24.84 \\
\bottomrule
\end{tabular}
}
\vspace{5pt}
\caption{\textbf{Running time for modules (milliseconds) on DTU~\cite{dtu}.}}
\label{Tab:runtime_module}
\end{table}

\begin{table*}[htbp]
\centering
\resizebox{1\linewidth}{!}{
\begin{tabular}{@{}lcccccccccc@{}}
\toprule
\multirow{2}{*}{Method} & \multirow{2}{*}{Optimization}  & \multicolumn{3}{c}{Real Forward-facing~\cite{llff}} & \multicolumn{3}{c}{NeRF Synthetic~\cite{nerf}}& \multicolumn{3}{c}{Tanks and Temples~\cite{tanks}} \\ 
\cmidrule(lr){3-5}\cmidrule(lr){6-8}\cmidrule(lr){9-11}
& & PSNR $\uparrow$ & SSIM $\uparrow$ & LPIPS $\downarrow$ & PSNR $\uparrow$ & SSIM $\uparrow$ & LPIPS $\downarrow$ & PSNR $\uparrow$ & SSIM $\uparrow$ & LPIPS $\downarrow$ \\
\midrule
3D-GS$_{7k}$~\cite{3Dgaussians} & \multirow{6}{*}{Gaussians} & 22.15 & 0.808 & 0.243 & 32.15 & 0.971 & 0.048 & 20.13 & 0.778 & 0.319 \\
3D-GS$_{30k}$~\cite{3Dgaussians} & & 23.92 & 0.822 & 0.213 & 31.87 & 0.969 & 0.050 & 23.65 & 0.867 & 0.184 \\
MVSGaussian~\cite{mvsgaussian} & & 23.87 & 0.856 & 0.169 & 25.31 & 0.943 & 0.098 & 22.70 & 0.872 & 0.187 \\
MVSGaussian$_{ft}$~\cite{mvsgaussian} & & \cellcolor{softorange}{26.98} & \cellcolor{softorange}{0.912} & \cellcolor{softorange}{0.116} & \cellcolor{softorange}{32.20} & \cellcolor{softorange}{0.972} & \cellcolor{softorange}{0.043} & \cellcolor{softorange}{24.82} & \cellcolor{softorange}{0.895} & \cellcolor{softorange}{0.171} \\
Ours & & \cellcolor{softyellow}{23.97} & \cellcolor{softyellow}{0.857} & \cellcolor{softyellow}{0.167} & \cellcolor{softyellow}{26.14} & \cellcolor{softyellow}{0.946} & \cellcolor{softyellow}{0.079} & \cellcolor{softyellow}{23.30} & \cellcolor{softyellow}{0.879} & \cellcolor{softyellow}{0.178} \\
Ours$_{ft}$ & & \cellcolor{softred}{27.03} & \cellcolor{softred}{0.913} & \cellcolor{softred}{0.113} & \cellcolor{softred}{32.41} & \cellcolor{softred}{0.973} & \cellcolor{softred}{0.042} & \cellcolor{softred}{25.00} & \cellcolor{softred}{0.899} & \cellcolor{softred}{0.162} \\
\bottomrule
\end{tabular}}
\vspace{5pt}
\caption{\textbf{Quantitative results after per-scene optimization using 3 source views on Real Forward-facing~\cite{llff}, NeRF Synthetic~\cite{nerf}, and Tanks and Temples~\cite{tanks} datasets.} For clarity in comparison, the scores labeled as ``Ours” refer to the generalization setting in Table~\ref{Tab:gen_llff}, while ``MVSGaussian$_{ft}$" and ``Ours$_{ft}$” represent results obtained after per-scene optimization.}
\label{Tab:fine-tuning}
\end{table*}

\section{Resource Consumption}

As shown in Table~\ref{Tab:runtime_comparison}, from the perspective of FPS (Frames Per Second) and GPU memory usage, our method achieves a balanced performance among the evaluated approaches. Specifically, for 3-view inputs, our method requires 1.01 GB, which is slightly higher than MVSGaussian~\cite{mvsgaussian} (0.88 GB) but remains within a reasonable range. Under 2-view inputs, memory consumption reduces to 0.89 GB, close to MVSGaussian~\cite{mvsgaussian} 
(0.87 GB), showing efficient scaling when fewer views are involved. The FPS of our method is 14.57 under 3-view inputs, which is lower compared to MVPGS~\cite{mvpgs} and MVSGaussian~\cite{mvsgaussian}. This suggests a trade-off between computational cost and performance fidelity. For 2-view inputs, the FPS improves to 15.50, aligning with the general trend of reduced computational demands with fewer views.

While our FPS is not the highest, it maintains a practical runtime for high-quality view synthesis. The memory consumption is competitive, reflecting efficient resource utilization despite the advanced modules and superior performance metrics. The trade-off between runtime and synthesis quality makes our method suitable for scenarios where fidelity takes precedence over speed.

Details of the time consumption across different stages of our main modules are demonstrated in Table~\ref{Tab:runtime_module}.

\section{Results after Per-Scene Optimization}
\label{subsec:per-scene optimization results}
Consistent with~\cite{mvsgaussian}, our method supports fine-tuning for improving rendering performance. During the per-scene optimization phase, we strictly adhere to the optimization strategy and hyperparameter settings established in~\cite{mvsgaussian}. Initialization of the 3D Gaussian representation (3D-GS) is performed using COLMAP~\cite{colmap} to reconstruct the point cloud from the dataset.

\begin{figure*}[htb]
    \centering
    \includegraphics[width=0.98\linewidth]{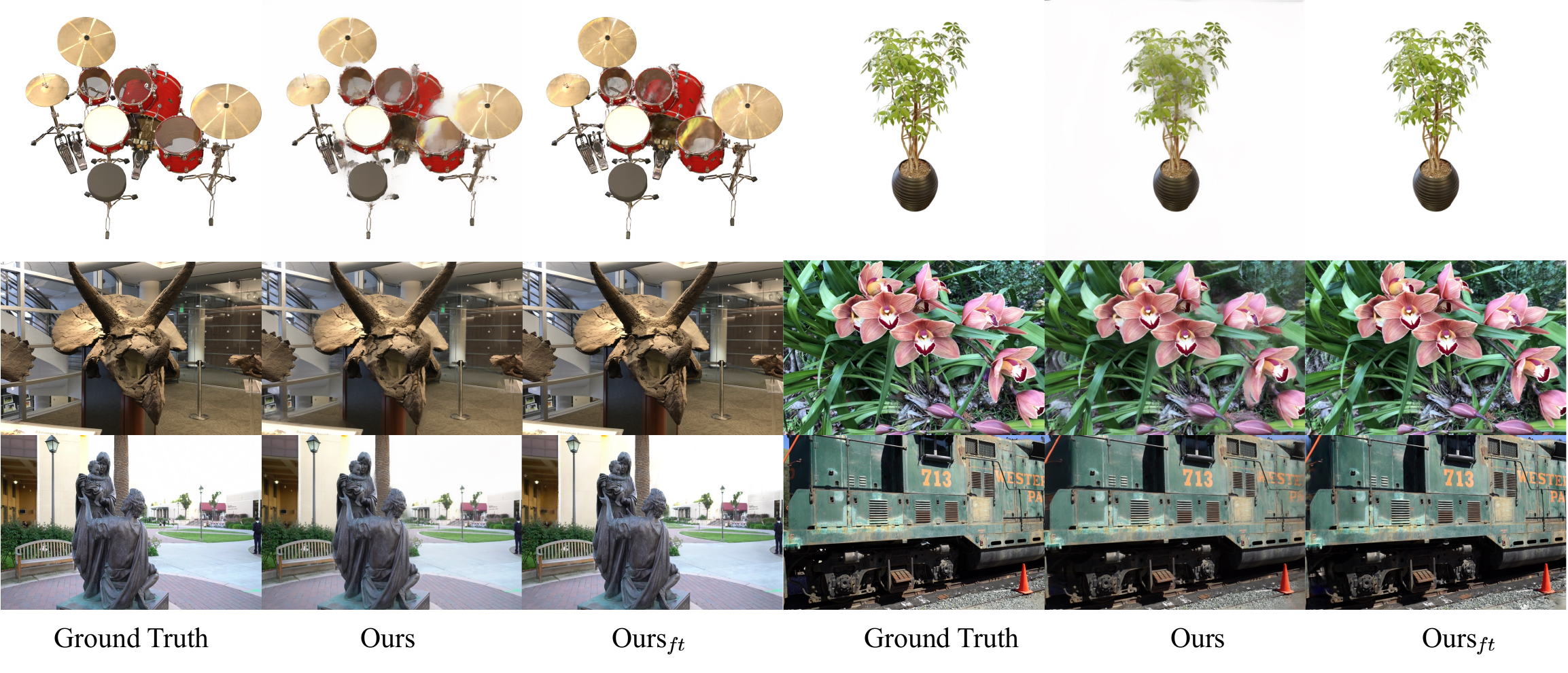}
    \vspace{5pt}
    \caption{\textbf{Qualitative comparison of rendered images using 3 source views on Real Forward-facing~\cite{llff}, NeRF Synthetic~\cite{nerf}, and Tanks and Temples~\cite{tanks} datasets.} ``Ours” refers to the generalization setting, while ``Ours$_{ft}$” represents results obtained after per-scene optimization. Our approach already produces good results in the generalizable setting, and for per-scene optimization, only a few additional iterations are required to achieve better rendering results.}
    \label{fig:vis_before_after_finetuning_supp}
\end{figure*}

The comparison of our method before and after optimization in the same scene is shown in Fig.~\ref{fig:vis_before_after_finetuning_supp}. Our approach already produces good results in the generalizable setting, and for per-scene optimization, only a few additional iterations are required to achieve better rendering results. 

\begin{figure*}[htb]
    \centering
    \includegraphics[width=0.98\linewidth]{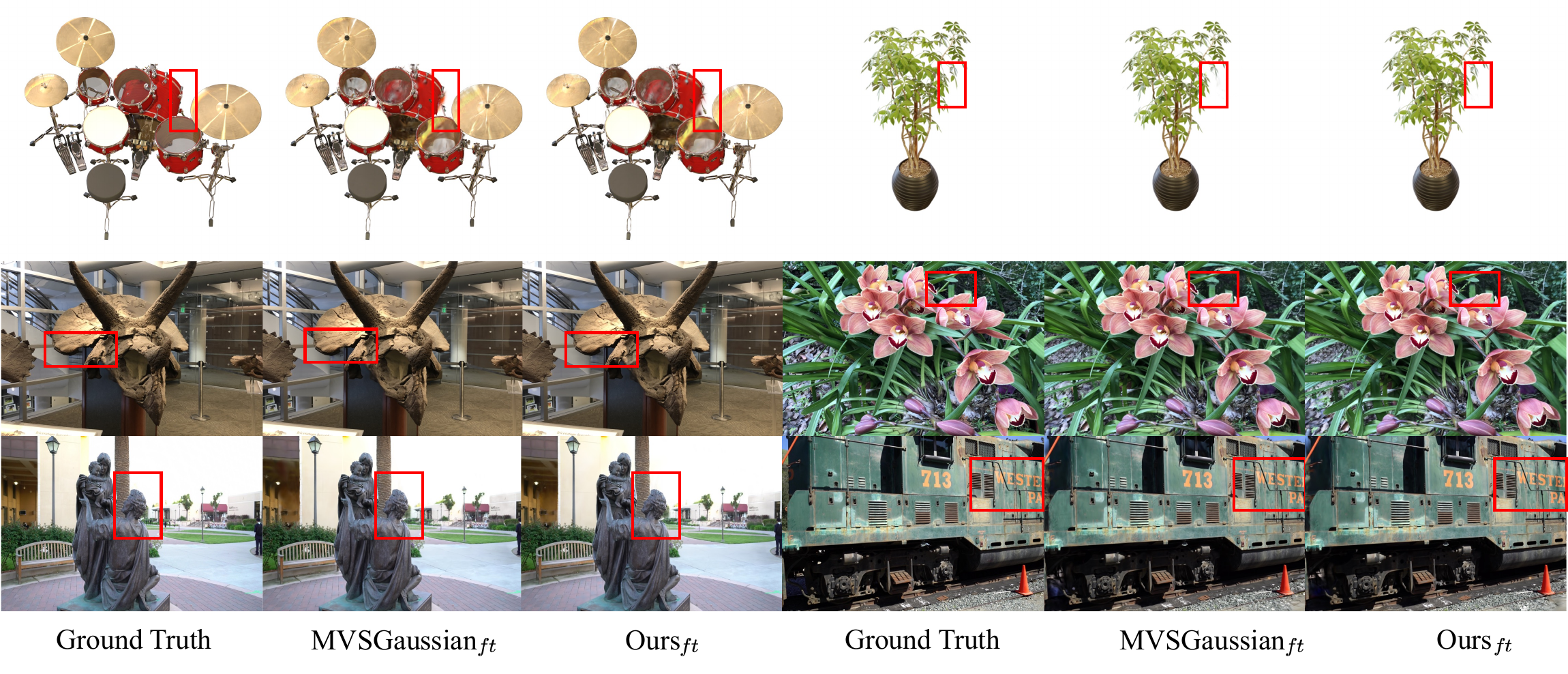}
    \vspace{5pt}
    \caption{\textbf{Qualitative comparison of rendered images using 3 source views on Real Forward-facing~\cite{llff}, NeRF Synthetic~\cite{nerf}, and Tanks and Temples~\cite{tanks} datasets after fine-tuning.} ``MVSGaussian$_{ft}$" and ``Ours$_{ft}$” represent results obtained after per-scene optimization.}
    \label{fig:vis_finetuning_supp}
\end{figure*}

We further compare rendered images with MVSGaussian~\cite{mvsgaussian}. After training scene by scene, MVGaussian~\cite{mvsgaussian} and our method can both achieve good results, because this is no longer a ``generalizable model". Even so, we still achieve better results. As depicted in Fig.~\ref{fig:vis_finetuning_supp}, the synthesized views retain intricate scene details and exhibit a marked reduction in artifacts compared to competing approaches. Leveraging the robust initialization from our generalizable model, our method achieves exceptional results with minimal fine-tuning. 

Quantitative results after per-scene optimization are presented in Table~\ref{Tab:fine-tuning}. We only optimize the Gaussians, significantly reducing both optimization time and rendering overhead. This efficiency is made possible by the strong initialization from our generalizable model and the robust design of our modules.

\section{More Visualizations}
\subsection{Visualization of Generalization Results}
\begin{figure*}[htb]
    \centering
    \includegraphics[width=0.98\linewidth]{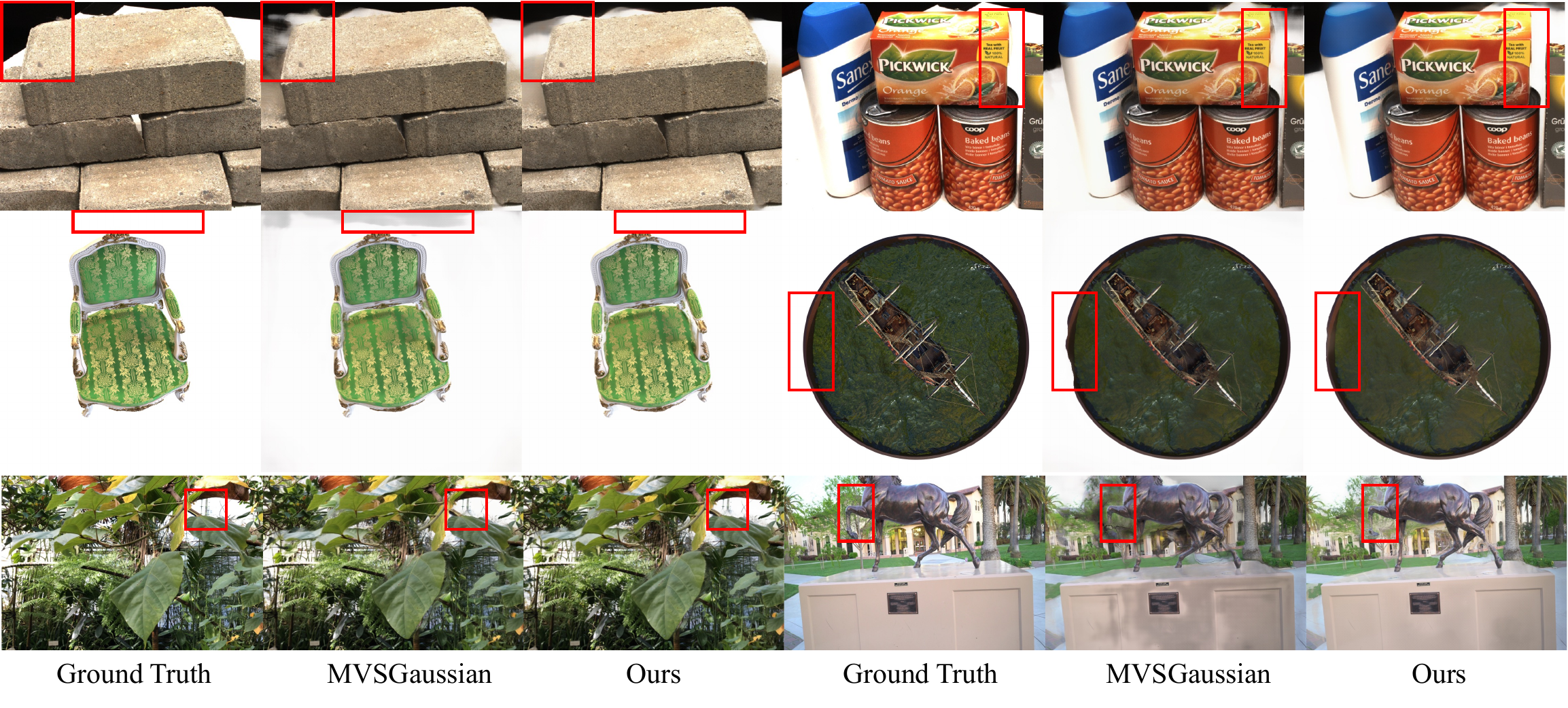}
    \vspace{5pt}
    \caption{\textbf{Qualitative comparison of rendered images using 3 source views on DTU~\cite{dtu}, Real Forward-facing~\cite{llff}, NeRF Synthetic~\cite{nerf}, and Tanks and Temples~\cite{tanks} datasets.} This is directly obtained using the generalizable model, the same as Fig~\ref{fig:vis_gen}.}
    \label{fig:vis_gen_supp}
\end{figure*}

Fig.~\ref{fig:vis_gen_supp} shows more examples of the same experimental setup as Fig.~\ref{fig:vis_gen}. It highlights a qualitative comparison of results under the generalization setting. Our approach demonstrates superior performance, particularly in challenging scenarios.

\subsection{Visualization of Depth Map}
\begin{figure*}[htb]
    \centering
    \includegraphics[width=0.98\linewidth]{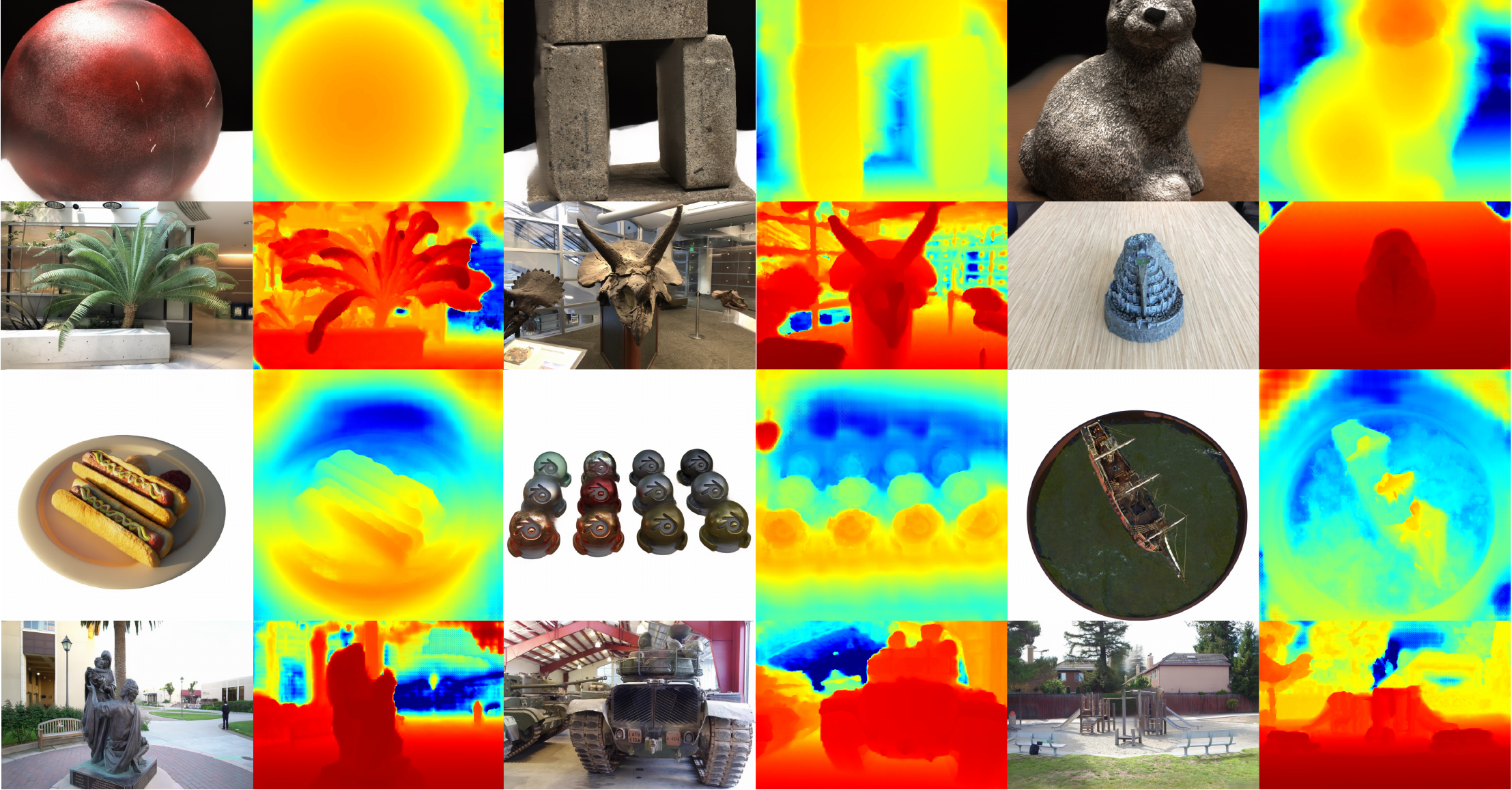}
    \vspace{5pt}
    \caption{\textbf{Depth maps visualization and corresponding rendered images on DTU~\cite{dtu}, Real Forward-facing~\cite{llff}, NeRF Synthetic~\cite{nerf}, and Tanks and Temples~\cite{tanks} datasets.} We can see that our depth map, as a good intermediate result, provides a solid foundation for the subsequent Gaussian expression.}
    \label{fig:depth_vis_supp}
\end{figure*}

Fig.~\ref{fig:depth_vis_supp} highlights the precision and smoothness of the depth maps generated by our approach, even in geometrically complex regions. For instance, our method produces sharp object boundaries and few artifacts, demonstrating its capability to capture fine-grained scene geometry. These results underscore the robustness of our approach in encoding spatially accurate and contextually aware features for novel view synthesis.

\subsection{Visualization of Source views}
\begin{figure*}[htb]
    \centering
    \includegraphics[width=0.98\linewidth]{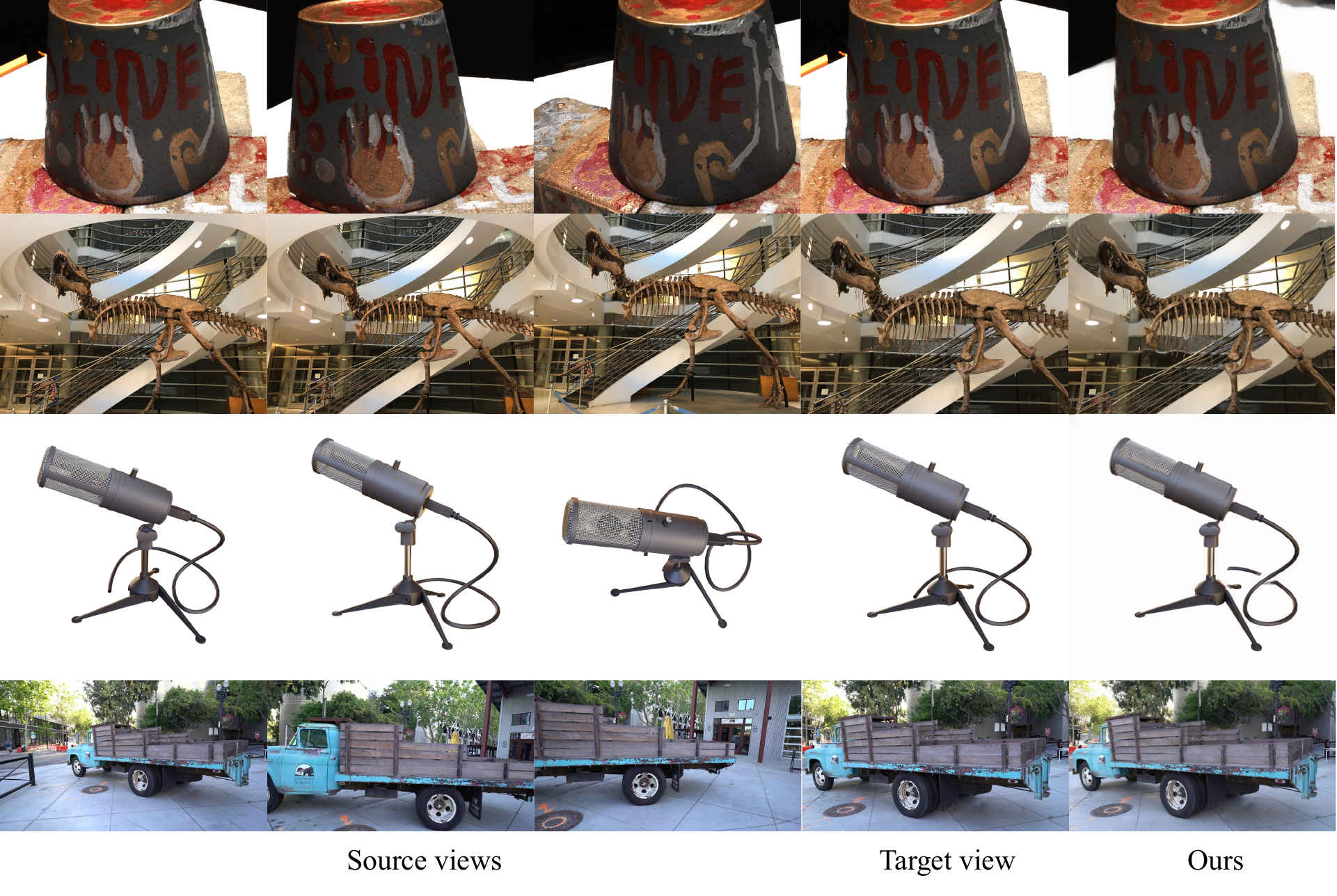}
    \vspace{5pt}
    \caption{\textbf{Source views visualization under the 3-view setting.} We select the three closest views as source views based on the distance of the viewpoints to render the target view. We also show the rendered images.}
    \label{fig:source_view_supp}
\end{figure*}

Fig.~\ref{fig:source_view_supp} shows the selection of our source views and their corresponding target views under the 3-view setting.

